\documentclass[11pt]{article}

\usepackage[preprint]{acl}

\usepackage{times}
\usepackage{latexsym}
\usepackage{amsmath}
\usepackage{listings}

\usepackage[T1]{fontenc}

\usepackage[utf8]{inputenc}
\usepackage[ruled, longend, linesnumbered]{algorithm2e}

\usepackage{microtype}

\usepackage{inconsolata}

\usepackage{graphicx}
\usepackage{multirow}
\usepackage{makecell}

\usepackage[most]{tcolorbox}
	
\definecolor{color_consultation}{RGB}{59, 162, 165}
\definecolor{color_configuration}{RGB}{194,169,211}
\definecolor{color_curation}{RGB}{255, 180, 0}
\title{MeNTi: Bridging Medical Calculator and LLM Agent\\with Nested Tool Calling}

\author{
 \textbf{Yakun Zhu\textsuperscript{1}},
 \textbf{Shaohang Wei\textsuperscript{2}},
 \textbf{Xu Wang\textsuperscript{1}},
 \textbf{Kui Xue\textsuperscript{3}},
\\
 \textbf{Shaoting Zhang\textsuperscript{3,4}},
 \textbf{Xiaofan Zhang\textsuperscript{1,3}}
\\
 \textsuperscript{1}Shanghai Jiao Tong University,
 \textsuperscript{2}Peking University,
\\
 \textsuperscript{3}Shanghai Artificial Intelligence Laboratory,
 \textsuperscript{4}SenseTime Research
\\
 \small{
   \textbf{Correspondence:} 
   \href{sjtuzykun@sjtu.edu.cn}{sjtuzykun@sjtu.edu.cn},
   \href{xiaofan.zhang@sjtu.edu.cn}{xiaofan.zhang@sjtu.edu.cn}
 }
}

\begin{document}

\maketitle
\begin{abstract}
Integrating tools into Large Language Models (LLMs) has facilitated the widespread application. Despite this, in specialized downstream task contexts, reliance solely on tools is insufficient to fully address the complexities of the real world. This particularly restricts the effective deployment of LLMs in fields such as medicine.
In this paper, we focus on the downstream tasks of medical calculators, which use standardized tests to assess an individual's health status.
We introduce MeNTi, a universal agent architecture for LLMs. MeNTi integrates a specialized medical toolkit and employs meta-tool and nested calling mechanisms to enhance LLM tool utilization. Specifically, it achieves flexible tool selection and nested tool calling to address practical issues faced in intricate medical scenarios, including calculator selection, slot filling, and unit conversion.
To assess the capabilities of LLMs for quantitative assessment throughout the clinical process of calculator scenarios, we introduce CalcQA. This benchmark requires LLMs to use medical calculators to perform calculations and assess patient health status. CalcQA is constructed by professional physicians and includes 100 case-calculator pairs, complemented by a toolkit of 281 medical tools.
The experimental results demonstrate significant performance improvements with our framework. This research paves new directions for applying LLMs in demanding scenarios of medicine\footnote{Code and Dataset available in \url{https://github.com/shzyk/MENTI} .}.
\end{abstract}

\section{Introduction}
\begin{figure}[ht!]
  \includegraphics[width=\linewidth]{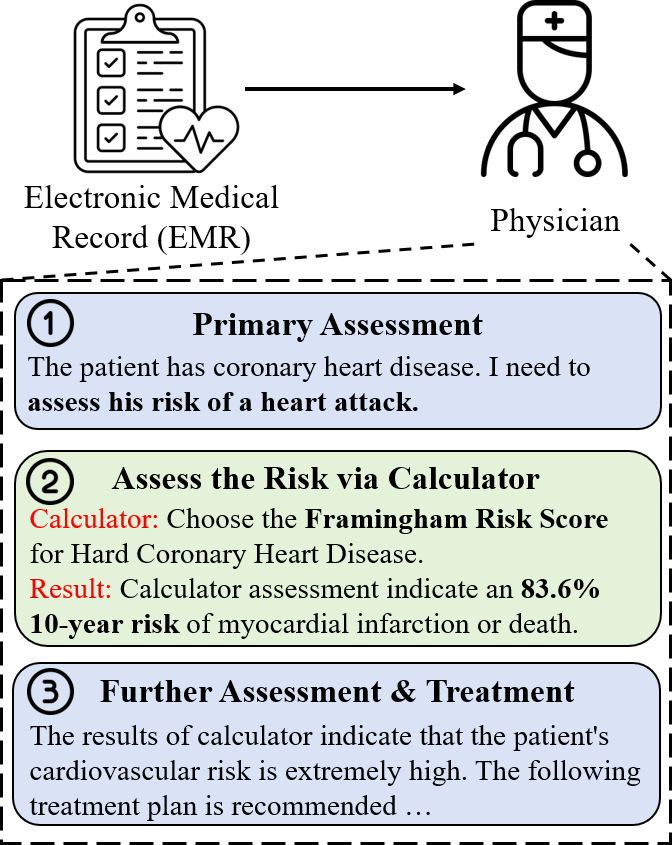}
  \caption{Instance of calculator application in Medical Scenarios. Physicians use calculators to assist in the quantitative assessment of patient's status based on their primary assessment.}
  \label{fig: calculator_instance}
  \vspace{-.3cm}
\end{figure}

Large Language Models (LLMs) have achieved milestone successes, demonstrating exceptional multi-tasking potential that includes reasoning, planning, tool usage, and code generation (\citealp{touvron2023llama}; \citealp{chenprogram}; \citealp{achiam2023gpt}; \citealp{ouyang2022training}). Recent studies indicate that by deeply integrating LLMs with large-scale toolsets, their task-solving efficiency can be significantly enhanced, further expanding the functional boundaries of the LLMs (\citealp{song2023restgpt}; \citealp{qin2023toolllm}; \citealp{gao2023pal}; \citealp{hao2024toolkengpt}). Additionally, developing customized tools tailored to specific application scenarios has become a key approach to optimize performance and enhance adaptability, showing remarkable results in numerous downstream tasks (\citealp{qian2023creator}; \citealp{cai2023large}; \citealp{yuan2023craft}).

In this paper, we focus on the issue of medical calculators, a typical and representative downstream task for LLMs tool application. Medical calculators are standardized tools used to quantitatively assess individual health status, functional levels, disease severity, and treatment outcomes. They are extensively applied in clinical diagnostics, therapeutic monitoring, and research fields, aiding medical professionals in precisely evaluating and guiding personalized medical decisions (\citealp{DZIADZKO20161}; \citealp{GREEN2019105002}; \citealp{dziadzko2016clinical}). Instances of calculator applications in medical scenarios are shown in Figure \ref{fig: calculator_instance}.
Calculators can help physicians quantify a patient's condition, providing a basis for formulating the next steps in the treatment plan.

However, given the complexity of medical calculators in practical applications, the current methods of tool application are insufficient. 
Specifically, selecting the appropriate tool is challenging. Firstly, there are currently over 700 calculators in use \footnote{Counted by https://www.medcentral.com/calculators/all}, and this vast number makes it difficult to choose the appropriate one. Secondly, each calculator is developed for different conditions, and the medical knowledge associated with these conditions, as well as the specific rules embedded within each calculator, make it hard to master them. Lastly, the continuous publication of new calculators every year adds ongoing pressure to keep tools up-to-date.
Consequently, we introduce the \textbf{meta-tool mechanism}, aimed at promoting flexible tool selection in vast toolkit. The meta-tool standardizes the timing of tool usage in medical calculator scenarios and refines the tool selection process, specifically catering to the selection within large-scale toolkits and complex scenarios.

Moreover, after selecting the tools, their application proves to be challenging due to locating information within lengthy case histories, slots filling, and unit conversion. Medical scenarios involve complex case histories, and selecting values from these contexts tests the capacity for long-context processing. Furthermore, the process demands substantial medical expertise to avoid errors in slot filling and to handle the challenges of unit conversion. For instance, LLMs may conflate rheumatic heart disease with congestive heart failure, or require conversions between units of total cholesterol from 8.3 mmol/L to 320.92 mg/dL. 
Consequently, we introduce the \textbf{nested calling mechanism}. Nested tool calling allows LLM to introduce additional tools when the current tools and information are insufficient for the task at hand. This mechanism is particularly valuable in scenarios requiring extra unit conversion tools, ensuring robust handling of complex medical calculation tasks.

We introduce a new benchmark, CalcQA, to validate the practicality of the llm agent throughout the entire clinical process of medical calculator scenarios. In collaboration with board-certified physicians, we develop this benchmark: physicians select calculators based on real patient cases and provide corresponding diagnostic pathways, then use GPT-4 to simulate dialogue scenarios.
Alongside this benchmark, we develop a specialized medical calculator toolkit. We compile 44 medical calculators that are widely used in medical practice and systematically organize conversion methods for 237 common medical units.

Our contributions are as follows: (1) We introduce the CalcQA benchmark, a new benchmark to assess the capabilities of LLMs in clinical calculator scenarios, including 100 calculator pairs based on real patients' cases, along with 281 medical calculator tools. (2) We develop the generalized MeNTi agent architecture, which expands the LLMs' ability to address real-world medical calculator tasks through meta-tool and nested calling mechanisms. (3) Our research indicates that MeNTi represents the first full-process implementation of LLMs for calculator assessments in real medical scenarios, demonstrating exceptional performance in executing such tasks.

\begin{figure*}[ht!]
  \includegraphics[width=\linewidth]{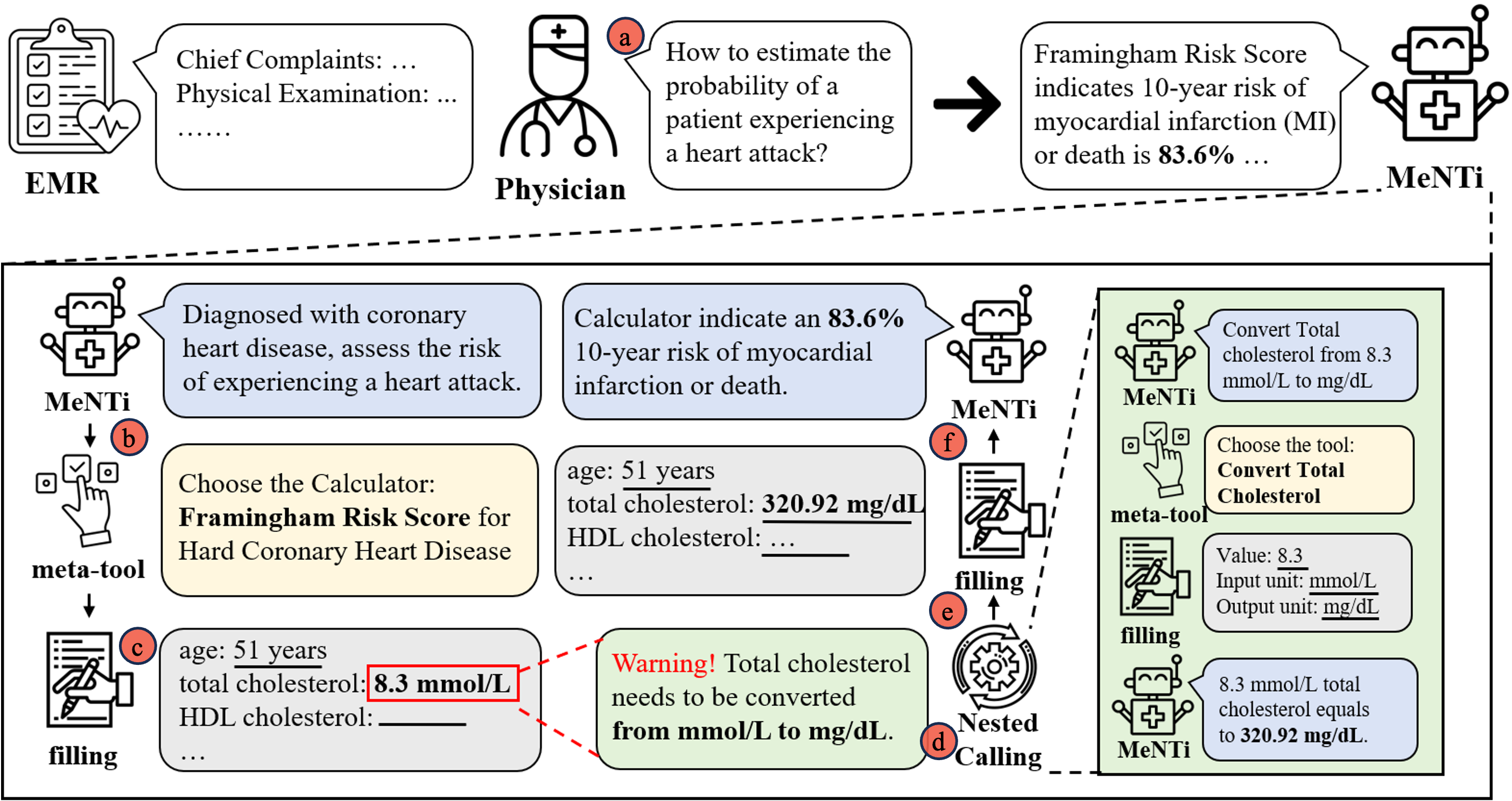}
  \caption{The whole workflow of MeNTi: (a) The physician proposes the next step for a more quantitative analysis. (b) Upon receiving the doctor’s request, MeNTi uses the meta-tool to determine and select the appropriate tool from the toolkit. (c) While filling in the parameters, MeNTi detects a unit mismatch and uses the nested tool calling mechanism to identify the need for additional tools. (d) MeNTi solves problems with more tools. (e) MeNTi consolidates the results from all tools and refills the parameters. (f) The final result is computed.}
  \label{fig: inference}
  \vspace{-.3cm}
\end{figure*}

\section{Related Works}

\textbf{LLMs for Medical Domain.}
With the application of LLMs across various fields, their penetration into the medical domain has become a hot research topic (\citealp{thirunavukarasu2023large}; \citealp{singhal2023large}; \citealp{clusmann2023future}). Recent studies focus on fine-tuning LLMs using real or synthetic medical data (\citealp{zeng2020meddialog}; \citealp{zhang2023deep}; \citealp{tang2023medagents}) to optimize their performance. For instance, PMC-LLaMA (\citealp{wu2024pmc}) has undergone pre-training by incorporating a vast resource of 4.9 million medical literature records. Similarly, ChatDoctor (\citealp{yunxiang2023chatdoctor}) integrates real doctor-patient communication data, enhancing the model’s ability to understand patient needs and make appropriate recommendations. Additionally, several studies have explored RAG to boost the efficacy of LLMs. The LLM-AMT (\citealp{wang2023augmenting}) adopts the RAG architecture, incorporating authoritative medical textbooks into the model. Self-BioRAG (\citealp{jeong2024improving}) trained with domain-specific retrievers, document databases, and instruction sets. Despite significant breakthroughs in the medical field, LLMs still face challenges in clinical calculator scenarios, due to inherent shortcomings in mathematical computation and logical reasoning. To further evaluate the capabilities of LLMs in calculator tasks, we introduce CalcQA, a comprehensive benchmark derived from real patients' cases.

\textbf{Agent and Tool Utilization.}
Although LLMs excel at information processing and linguistic interaction, they cannot directly execute specific tasks such as computation and calendar (\citealp{chen2024llmarena}; \citealp{huang2023metatool}; \citealp{ruan2023tptu}). Recent research focuses on integrating LLMs with real-world APIs to enhance their practical capabilities (\citealp{wang2024executable}; \citealp{tang2023toolalpaca}; \citealp{schick2024toolformer}). This integration generally involves three stages: tool creation, selection, and application. Regarding tool creation, previous practices have relied on existing APIs, which present limitations in specialized fields. LATM (\citealp{cai2023large}) has pioneered creating tools by generating code. CRAFT (\citealp{yuan2023craft}) focuses on learning for datasets to create tools. Regarding tool selection, the method of instructing LLMs with prompts (\citealp{zhuang2024toolqa}) is suitable for environments with smaller tool libraries, while the technique of using tool names for dense retrieval (\citealp{qin2023toolllm}) offers another approach. As for tool application, the AnyTool (\citealp{du2024anytool}) introduces self-reflection to guide operational processes. However, given the complexity of downstream tasks in the medical field, there is still a lack of an agent framework capable of flexible tool selection and nested tool calling. This limitation restricts the efficacy of LLMs in complex scenarios.

\section{Methodology}
In Section \ref{sec: benchmark}, we introduce CalcQA, a novel benchmark and a toolkit with specialized medical tools, to assess the ability of LLMs to perform end-to-end calculator assessment in clinical scenarios. We develop this benchmark by having professional physicians diagnose real cases and then structuring these cases and diagnostic results into question-answer pairs.
In Sections \ref{sec: curation} and \ref{sec: nested}, we introduce MeNTi, which employs the meta-tool for tool selection and nested calling mechanism to assist LLMs in tool utilization. By enabling flexible tool selection and nested tool calling, MeNTi effectively addresses intricate scenario tasks with the specialized toolkit.

Specifically, after the physician conducts a primary assessment of the patient and identifies the next steps for quantification, the task is handed over to MeNTi. After preliminary diagnosis, MeNTi uses the meta-tool (Section \ref{sec: curation}) to select an appropriate medical calculator. MeNTi then proceeds to fill in the slots of calculator parameters. If MeNTi detects a mismatch in parameter units, it employs the nested tool calling (Section \ref{sec: nested}). This mechanism enables iterative tool selection and utilization to resolve unit mismatches caused by complex toolchains requiring multiple tools. Once the conversion tool's results are integrated, MeNTi continues filling the calculator parameters and completes the final computation. This workflow, illustrated with a concrete example in Figure \ref{fig: inference}, is further detailed with complete patient information and case examples in the appendix \ref{apdx: example}.

\subsection{Benchmark Creation}\label{sec: benchmark}
Existing LLM benchmarks for medical calculators primarily focus on assessing the LLMs’ ability to handle medical knowledge-based questions or their capacity to make selective diagnostic judgments in specific scenarios. However, these benchmarks are not designed directly from clinical practice and do not derive diagnostic conclusions directly from actual cases, making them insufficient for evaluating the usability of LLMs in real clinical settings. Therefore, we introduce CalcQA, a new benchmark constructed based on real cases, aimed at assessing LLMs within the context of medical calculators.

\begin{figure}[t]
  \includegraphics[width=\linewidth]{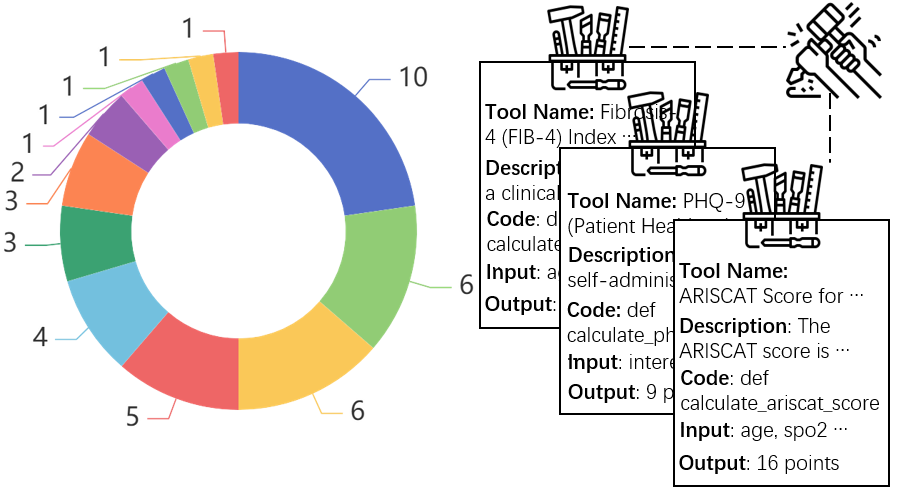}
  \caption{Distribution map of the assessment calculator departments. The calculator used in this study encompasses 13 clinical departments, with a particular emphasis on cardiology, involving 11 calculators.}
  \label{fig: calculator_class}
  \vspace{-.3cm}
\end{figure}

\textbf{Toolkit Construction.}
Given the specialized medical knowledge and complex computational logic involved in medical calculators, LLMs require external knowledge. Accordingly, we develop a generalized medical toolkit (\citealp{cai2023large}; \citealp{yuan2023craft}; \citealp{yuan2024easytool}).
Initially, we scrap essential knowledge from authoritative medical websites, then utilized GPT-4 for code generation to perform precise mathematical calculations. Finally, we design real-world scenarios to verify our tools. After meticulous selection and integration, we compile 44 medical calculators that are widely used in medical practice, whose application scenarios and precise calculation rules are also represented. These calculators extensively cover multiple branches of medicine, as detailed in Figure \ref{fig: calculator_class}. Additionally, considering that medical calculations often involve precise conversions between medical units, like conversion between the mass and molar quantities, we organize conversion tools for 237 common medical units. The manufacturing process and content are consistent with those of the calculators above.

\textbf{Annotation.}
In this benchmark, each question is derived from actual clinical cases and diagnosed by professional physicians to serve as the ground truth.
We initially employed a stratified sampling strategy based on the distribution of selected tools, extracting 100 cases proportionally from different clinical departments. Subsequently, we hired experienced physicians to select the appropriate calculators for each case and to thoroughly document the rationale for their choices, their diagnostic opinions, and calculator assessment results. On this basis, we utilized LLMs in conjunction with detailed diagnostic information to generate restrained user queries, thereby creating the CalcQA benchmark. This benchmark includes specific user queries, case details, and diagnostic information, with expected answers being the names of the recommended calculators and their computed results.

\subsection{Meta-Tool for Tool Selection}
\label{sec: curation}

Existing tool selection strategies often rely on presenting a complete toolkit to LLMs, expecting them to autonomously choose the appropriate ones. However, this approach is impractical for large toolsets. Relying solely on retrieval methods to match tools with specific task scenarios is also challenging, which requires deep insights into the nature of the tasks and a thorough understanding of tools.
Additionally, the existing timing strategy for applying tools is determined by fine-tuned LLMs. Although this approach has widespread applicability, its inherent high level of uncertainty is unacceptable in medical environments that prioritize precision and reliability.
Therefore, we introduce the meta-tool mechanism specifically to address tool selection challenges. 

The meta-tool is tailored for medical calculator scenarios, standardizing the timing of tool usage based on specific situations, thus obviating the need for LLMs of tool calling capabilities. Specifically, tool calling occurs in two scenarios: first, when the user expresses the need for calculator assessment during the inquiry process; second, when the current task scenario lacks the capabilities required to resolve the task, necessitating the introduction of additional tools to augment existing abilities. In the latter case, the nested calling (\ref{sec: nested}) presents specific tool usage requirements to the meta-tool, which then assesses the demands for tool application and proceeds to match the appropriate tools based on the identified requirements.


\begin{algorithm}
\KwData{Toolkit $T=\{ t_c^i\}$, where $c$ is tool category and $i$ is tool index; User Query $Q$, Case Diagnosis $D$, Patient Case History $H$, Prompt $P$, Similarity Ranking $R$; LLM Model $\mathrm{F}$, Retrieval Model $\mathrm{E}$, Rerank Method $\mathrm{RRF}$.}
\KwResult{Selected tool by meta-tool, $S$.}
\tcc{Meta-tool selects tools based on the query and diagnosis.}
\Begin{
    $D \gets \mathrm{F}(H \mid P_{diagnosis})$\;
    $c_{classfier} \gets \mathrm{F}(Q, D \mid P_{classfier})$\;
    $Q_{rewriter} \gets \mathrm{F}(Q, D \mid P_{rewriter})$\;
    $\{R^q_{key}\} \gets \mathrm{E}(Q_{rewriter}, \{t^i_{c_{classfier}}\})$\;
    $R_{topk} \gets \mathrm{RRF} (\{R^q_{key}\})$\;
    $S \gets \mathrm{F} (Q, H, R_{topk} \mid P_{dispathcer})$\;
}
\caption{Meta-Tool Selection Process}
\label{algorithm}
\end{algorithm}

\begin{figure}[t]
  \includegraphics[width=\columnwidth]{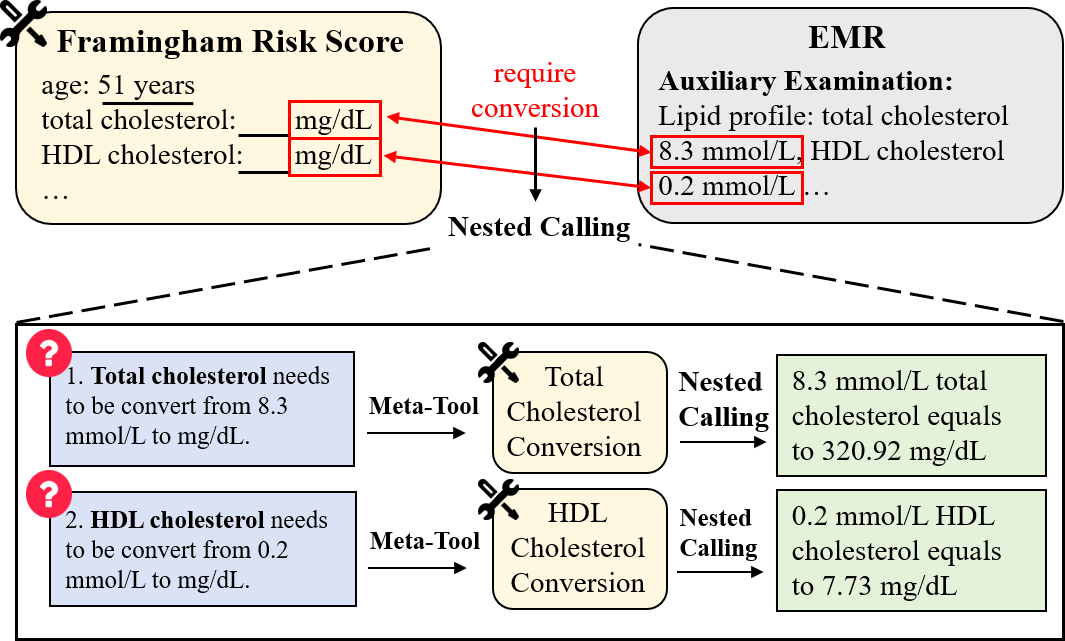}
  \caption{Flow of Nested Tool Calling. MeNTi enhances LLMs' tool utilization capabilities using meta-tool and nested calling mechanism, while also extending the knowledge boundaries of LLMs with a specialized medical toolkit, complementing each other. Through the nested tool calling of the toolkit, MeNTi provides LLMs with more operation and information for solving practical tasks.}
  \label{fig: tool_calling}
  \vspace{-.3cm}
\end{figure}

Additionally, the meta-tool facilitates a flexible tool selection that aligns with various scenarios, assisting the MeNTi system in choosing appropriate tools to complete real tasks.
The meta-tool achieves tool selection through the following steps, as it is shown in Algorithm \ref{algorithm}.
(1) Calculator Classifier. In MeNTi, tools are categorized based on their origin and application scenarios (\citealp{du2024anytool}; \citealp{song2023restgpt}), like scale tools for assessment and unit tools for conversion. The classifier instructs the LLM to accomplish tool-type localization by matching the requirements of the task.
(2) Query Rewriter. The selection of calculators needs to be based on the patient's condition, and Rewriter integrates the user's query intent with the diagnosis overview, generating insightful query proposals (\citealp{wang2023query2doc}; \citealp{gao2023precise}; \citealp{jagerman2023query}), denoted as $Q$. The diagnostic overview is provided by the LLMs in conjunction with the patient's condition.
(3) Tool Retrieval \& Reranker. The retrieval uses the query $q$ and employs the dense retrieval method (\citealp{zhoudocprompting}; \citealp{lewis2020retrieval}; \citealp{yuan2023craft}). This component also utilizes a multi-key strategy to enhance the precision, the tool’s name $n$, description $d$, and the docstring associated with the tool $s$. By calculating the similarity with the query $q$, three different similarity rankings of $R^q$ are constructed: name-based $R^q_{n}$, a fusion of name and description $R^q_{n+d}$, and a fusion of name and docstring $R^q_{n+s}$. That is $\forall  q \in Q$: 
\begin{equation*}
\begin{aligned}
    R^q_{n} & = \operatorname{ArgMax}(\operatorname{sim} (n, q) | N \subset T)\quad \\
    R^q_{n+d} & = \operatorname{ArgMax}(\operatorname{sim} (n+d, q) | N, D \subset T)\quad \\
    R^q_{n+s} & = \operatorname{ArgMax}(\operatorname{sim} (n+s, q) | N, S \subset T)\quad
\end{aligned}
\end{equation*}
Then, the reranker adopts the Reciprocal Rank Fusion (\citealp{cormack2009reciprocal}), applied to reranking in multiple retrievals. Each query utilizes the rewriter ($|Q|$) and multiple keywords ($|R^q|$) to perform multiple retrieval operations, to select tools that are robust and superior across scenarios.
(4) Dispatch tool in actual scenario. Retrieval based on semantic similarity matching does not necessarily ensure that tools are well-suited to the actual scenarios. The dispatcher integrates background information about the task and detailed descriptions of each candidate tool, instructing the LLM to consider both comprehensively (\citealp{shen2024hugginggpt}; \citealp{zhao2024expel}).

\subsection{Nested Tool Calling}
\label{sec: nested}

Previous studies primarily focus on aggregating more tools to expand the functional boundaries of LLMs. However, the application scenarios for LLMs are far from straightforward, especially in medical calculator tasks. Specifically, once a suitable calculator is chosen, its application presents significant challenges. Medical scenarios involve complex case histories, and selecting parameters from these contexts tests the capacity for long-context processing. Furthermore, the process demands substantial medical expertise to avoid errors in parameter judgment and to handle the challenges of unit conversion. This demands the appropriate use of current tools according to the context and the ability to identify unmet demands and integrate different tools, which is essentially the capability of nested tool calling.

Therefore, we introduce the nested calling mechanism which consists of three steps: First, it involves the tool slot filling from the patient's case history and previous tool outcomes; second, it verifies whether the current task requirements have been met; and third, if not, the nested calling mechanism plans subsequent demands for the agent, who then continues to orchestrate various tools for nested tool calling, ensuring the final task requirements are satisfied.
In the first step, we facilitate the slot filling by inputting comprehensive medical records, tool descriptions, and their function docstrings into the LLM, which encompasses the specific functionalities of the tools, their expected outputs, parameter constraints, and more, enabling accurate information choices, and then to use the tool appropriately.

In subsequent steps, the nested calling reviews the tool slots filling, ensuring that they meet the prerequisites of each tool application. If there are obstacles in the filling and the task requirements, the nested calling records the reasons for failure, and through natural language analysis troubleshooting solutions and subsequent requirements.
The subsequent requirements are conveyed to the MeNTi agent. MeNTi continues to select more tools by meta-tool to provide additional information or handle tasks to assist in resolving the current task. During this process, MeNTi continues to utilize the nested calling to facilitate tool integration. This nested calling leverages planning capabilities based on the current task state, and reusing tools within MeNTi, enables MeNTi to demonstrate a high degree of flexibility and adaptability in handling complex task scenarios. This workflow is depicted in Figure \ref{fig: tool_calling}.

\section{Experiment}

\begin{table*}[ht!]
  \centering
  \begin{tabular}{cclc|cccc}
    \hline
    \textbf{Category}& \textbf{Method}& \textbf{Model} &\textbf{Param. Size} &\textbf{CSA} & \textbf{SFA} & \textbf{UCA} &\textbf{CCA}\\
    \hline
     \multirow{5}{*}{Basic Reasoning}& \multirow{5}{*}{CoT} &BianQue2& 6B & 0\% & 0\% & 0\% &0\%\\
     &&ChatGLM3 & 6B & 31\% & 0\% & 0\%& 0\%\\
     &&LLaMA-3.1 & 8B &65\%	&33.4\%	&0\% &13\% \\
     &&PULSE  & 20B & 79\% & 11.2\% & 0\% & 7\%\\
     &&GPT-3.5 Turbo & / & 74\% & 60.4\% & 4.1\% & 10\%\\
     &&GPT-4o & / & 86\% & \textbf{75.7\%} & 22.4\% & 22\%\\
    \hline
    \multirow{3}{*}{Tool Application}&\multirow{3}{*}{CRAFT}& PULSE & 20B & 0\% & 0\% & 0\% & 0\%\\
    && GPT-3.5 Turbo & / & 25\% & 4.1\% & 0\% & 9\%\\
    && GPT-4o & / & 28\% & 4.9\% & 0\% & 19\%\\
    \hline
    \multirow{3}{*}{Our Framework}& \multirow{3}{*}{MeNTi}& PULSE & 20B & 81\% & 35.2\% & 46.9\%& 36\%\\
    && GPT-3.5 Turbo & / & 77\% & 56.2\% & 20.5\% & 24\%\\
    && GPT-4o & / & \textbf{86\%} & 70.3\% & \textbf{69.4\%} & \textbf{49\%}\\
    \hline
  \end{tabular}
  \caption{The experimental results of MeNTi and other categories of baselines on CalcQA.}
  \label{table: main_result}
  \vspace{-.5cm}
\end{table*}


In this section, we test the performance of MeNTi on CalcQA and compare it with other methods (\ref{sec: exp_set} \& \ref{sec: exp_res}). Additionally, we conduct further analysis (\ref{sec: ablation}), showing the rational design of MeNTi.
\subsection{Experimental Setting}
\label{sec: exp_set}

\textbf{Metrics.}
We employ four novel metrics, designed to quantify the accuracy of agents in selecting appropriate calculators, the agents' ability to fill tool slots, their competence for nested tool calling, and the overall performance of MeNTi in executing end-to-end tasks.

\begin{itemize}
    \item \textbf{Calculator Selection Accuracy (CSA)}: We calculate the percentage of cases where the agent's choice of calculators aligned with the ground truth. This metric measures the agent's ability to select tools based on user demands and patient conditions correctly.
    \item \textbf{Slot Filling Accuracy (SFA)}: We calculate the percentage of slots where the agents correctly fill it. This metric assesses the agent's ability to accurately fill in the slots of medical calculators.
    \item \textbf{Unit Converting Accuracy (UCA)}: 
    We calculate the percentage of slots that require unit conversions and the agent correctly fills. This metric measures the agent's ability to handle complex medical calculator slots.
    \item \textbf{Calculator Calculation Accuracy (CCA)}: We calculate the percentage of cases in which the specific calculator assessment values provided by the agent matched the ground truth. This metric measures the agent's capability to execute complete end-to-end tasks within medical calculator scenarios.
\end{itemize}

\textbf{Baselines.} We compare the MeNTi with 2 kinds of methods on our benchmark.
\begin{itemize}
    \item \textbf{Basic Reasoning:} We explore the capabilities of medical LLMs, directly tackling the task to see if they can select the appropriate calculators and provide evaluation results without the use of tools. We use the CoT approach. Three open-source medical models are selected: BianQue2 (\citealp{chen2023bianque}), ChatGLM3 (\citealp{du2021glm}), and PULSE (\citealp{pulse2023}). Additionally, LLaMA-3.1, GPT-3.5-Turbo, and GPT-4o are chosen for comparison.
    \item \textbf{Tool Application:} We adopt the tool application framework CRAFT (\citealp{yuan2023craft}) to explore its capabilities in executing complex tool operations. To ensure a fair evaluation, the customized toolkit is provided. To facilitate comparison with the MeNTi framework, we employed PULSE, GPT-3.5-Turbo, and GPT-4o as the backbone.
    \item \textbf{Our Framework:} We employed multiple LLMs to drive testing of the MeNTi framework. The objective is to evaluate the framework's performance across different backbones, aiming to verify its robust generalization capabilities and adaptability. PULSE, GPT-3.5-Turbo, and GPT-4o are chosen.
\end{itemize}

\subsection{Main results}
\label{sec: exp_res}
The results are presented in Table \ref{table: main_result}. We observe significant performance degradation in some small models under the Basic Reasoning when processing long contexts, highlighting the inherent difficulties of real medical scenario tasks. In contrast, LLaMA-3.1, PULSE, GPT-3.5, and GPT-4o demonstrate outstanding performance in tool selection due to their superior model capabilities. However, they still feel short in completing unit conversion and task computations. While the CRAFT framework achieves some performance improvements in GPT-4o, the extent of improvement is limited, further confirming that simple tool application strategies are inadequate for addressing the complexities of real-world medical tasks.

Overall, MeNTi demonstrates exceptional performance across various backbones, with notable improvements in GPT-4o. Based on this, we derive the following five experimental conclusions. (1) Compared to traditional baselines, MeNTi demonstrates significant performance enhancements in both UCA and CCA. This validates the substantial improvements of our framework, particularly the nested calling mechanism, over existing methods. (2) Although the meta-tool shows limited improvement on the CSA, primarily due to the benchmark containing only 44 commonly used medical calculators, its application in 237 unit conversion tools significantly advances the UCA. This demonstrates the effectiveness of the meta-tool, especially when dealing with a larger toolkit. (3) The significant advancements in UCA appear to slightly undermine the SFA. This may be attributed to the introduction of unit conversion, which complicates slot filling and consequently affects its efficiency. (4) MeNTi achieves remarkable results in both medical LLM like PULSE and general LLM like GPT-3.5 and GPT-4o, demonstrating its high generalizability and broad applicability. (5) The outstanding performance of GPT-4o within the MeNTi further validates the potential of MeNTi in superior LLMs, suggesting that its performance ceiling is far from being reached.

It is noteworthy that in the calculation of the CCA index, given the rounding errors inherent in numerical operations, we acknowledge a tolerance range of $\pm 0.5$. Based on this, we achieve a result of 49\% in the GPT-4o model. When we expand this tolerance to $\pm 1.5$, the performance of the GPT-4o model improves to 68\%; further extending the tolerance to $\pm 2.5$ resulted in a performance of 77\%. These data indicate that, despite some inaccuracies, the evaluation results from MeNTi still hold significant referential value. Through case study analysis, we identify that these errors primarily stem from the Knowledge Hallucination presented by LLMs. For example, when using the "CHA2DS2-VASc Score for Atrial Fibrillation Stroke Risk" calculator, LLMs incorrectly categorize a patient’s history of rheumatic heart disease as congestive heart failure, though there is no direct equivalence between them. This knowledge hallucination leads to subtle errors in the parameter population, ultimately impacting the results.

\subsection{Further Analysis}
\label{sec: ablation}
To comprehensively evaluate the academic contribution of the MeNTi framework, we systematically implemented a series of studies. Results are shown in Table \ref{table: alternative} and Table \ref{table: ablation}.

\subsubsection{Alternative Component Analysis}
To further evaluate the performance improvement brought by MeNTi, in addition to using the LLM approach described in Section \ref{sec: exp_set}, we select more methods and conduct comparative experiments by replacing certain components of MeNTi:
(1) Alternative Retrieval. We compare our approach with the commonly used retrieval methods, BM25(\citealp{robertson2009probabilistic}), SimCSE(\citealp{gao2021simcse}) and M3E(\citealp{Moka}). We only replace the meta-tool for tool selection with these methods while keeping the toolkit unchanged.
(2) Alternative Calculator. We select ToRA (\citealp{gou2023tora}), a model for solving mathematical problems, to address the calculator challenges, contrasting it with our toolkit approach. We replace only the nested calling mechanism for calculation with ToRA.
\begin{table}[ht!]
  \centering
  \begin{tabular}{lc|cc}
    \hline
    \textbf{Model} &\textbf{Param. Size} &\textbf{CSA} &\textbf{CCA}\\
    \hline
    BM25& /&0\%& - \\
    SimCSE& /&2\%& - \\
    M3E& /&39\%& - \\
    \hline
    ToRA & 13B & - & 0\%\\
    ToRA-Code& 13B & - & 0\%\\
    \hline
    PULSE& 20B & 81\%&36\%\\
    GPT-3.5 Turbo & / & 77\%& 24\%\\
    GPT-4o & / & \textbf{86\%}& \textbf{49\%}\\
    \hline
  \end{tabular}
  \caption{The alternative component analysis of MeNTi.}
  \vspace{-.3cm}
  \label{table: alternative}
\end{table}

Overall, MeNTi demonstrates exceptional performance. Experimental data indicate that traditional retrieval methods are only moderately effective in tool selection, and purely semantic retrieval mechanisms lack reliability in real-world scenarios. While retrieval tools are utilized for handling large-scale datasets, the assistance of LLM is still necessary for optimal tool selection. Unfortunately, due to ToRA's limitations in processing long context, it is inadequate for handling tasks in medical scenarios involving long patient case histories. Therefore, employing the agent to enhance tools remains essential at present.

\subsubsection{Ablation Analysis}
We conducted ablation experiments on the tool selection process of the meta-tool. The meta-tool implements tool selection through four steps (shown as Algorithm \ref{algorithm}). The Classifier performs the initial classification of tools, the Rewriter explores and expands the selection requirements, and the Retrieval step uses the name, docstring, and description as keys for preliminary similarity-based tool retrieval. Finally, the Dispatcher completes the more granular tool selection in actual scenarios. We will sequentially ablate each component and evaluate its impact.

\textbf{Classifier.}  We discard the previous tool categorization methods and directly retrieve all tools for the experiment. The results demonstrate that the removal of the Classifier significantly reduces the CSA and UCA. This underscores the importance of tool management, particularly when dealing with a large-scale toolkit.

\begin{table}[ht!]
  \centering 
  \begin{tabular}{l|cccc}
    \hline
    \textbf{Framework} & \textbf{CSA} & \textbf{SFA}& \textbf{UCA}& \textbf{CCA}\\
    \hline
    MeNTi & 81\% & 35.2\% & 46.9\%& 36\%\\
    w/o Classifier & 57\%& 18.4\% & 10.2\%& 24\%\\
    w/o Rewriter & 64\% & 21.8\% & 26.5\%& 27\%\\
    w/o Ret. name& 30\%& 12.1\% & 26.5\%& 15\%\\
    w/o Ret. doc.& 34\%& 12.8\% & 26.5\%& 15\%\\
    w/o Ret. desc.& 31\%& 12.5\% & 28.5\%& 15\%\\
    w/o Dispatcher& 37\% & 11.9\% & 16.3\%& 22\%\\
    \hline
  \end{tabular}
  \caption{Ablation of MeNTi framework.}
  \vspace{-.3cm}
  \label{table: ablation}
\end{table}

\textbf{Rewriter.} Rewriter of the meta-tool combines user queries with patient conditions to uncover professional insights. We discard this component and use the original query for retrieval. The results show a moderate reduction, highlighting the importance of integrating medical calculator tasks with patient conditions.

\textbf{Retrieval.} Retrieval of the meta-tool is designed to perform preliminary tool filtering based on the semantics of the tools. Our Retrieval adopts a multi-key search strategy. By incrementally removing each keyword, we delve into the specific utility of each keyword. All three keys are important, as removing any of them significantly impacts performance.

\textbf{Dispatcher.} Dispatcher of the meta-tool integrates real-world task scenarios and tool application strategies to select the most suitable tools. We remove the Dispatcher and use the tool with the highest semantic similarity. The results indicate that in practical tool usage scenarios, it is essential to select tools based on their actual functionalities.

\section{Conclusion}
In this work, we introduce MeNTi, an agent framework that achieves flexible tool selection and nested tool calling, specifically designed for medical calculator scenarios. We develop and validate a medical calculator toolkit. Additionally, we introduced CalcQA, a benchmark consisting of 100 case-assessment pairs, to assess the capabilities of LLMs in medical calculator tasks. Extensive experiments have shown that our method achieves superior performance compared to existing methods. We hope our work will further inspire the application of LLMs in medical contexts.

\section*{Limitations}
Despite MeNTi making significant strides in medical calculator tasks through flexible tool selection and nested tool calling, it still faces several limitations that pose challenges to its overall performance. The primary issue is the inherent complexity and length of medical cases, which makes MeNTi's capabilities seem relatively insufficient on a backbone with weaker long-context abilities. Additionally, due to the high cost of data annotation in medical calculator scenarios, the current CalcQA benchmark only covers 100 instances, necessitating more examples to further enhance the model's generalization validation. Furthermore, although the MeNTi framework theoretically has broad applicability, the high difficulty and cost of tool development have prevented us from systematically evaluating its generalization capabilities across other downstream tasks, a shortfall that limits the exploration of its potential for wider application. These issues merit further research and consideration.

\bibliography{custom}

\appendix

\section{Implementation Details}\label{implement}
\begin{figure*}[t]
  \includegraphics[width=\linewidth]{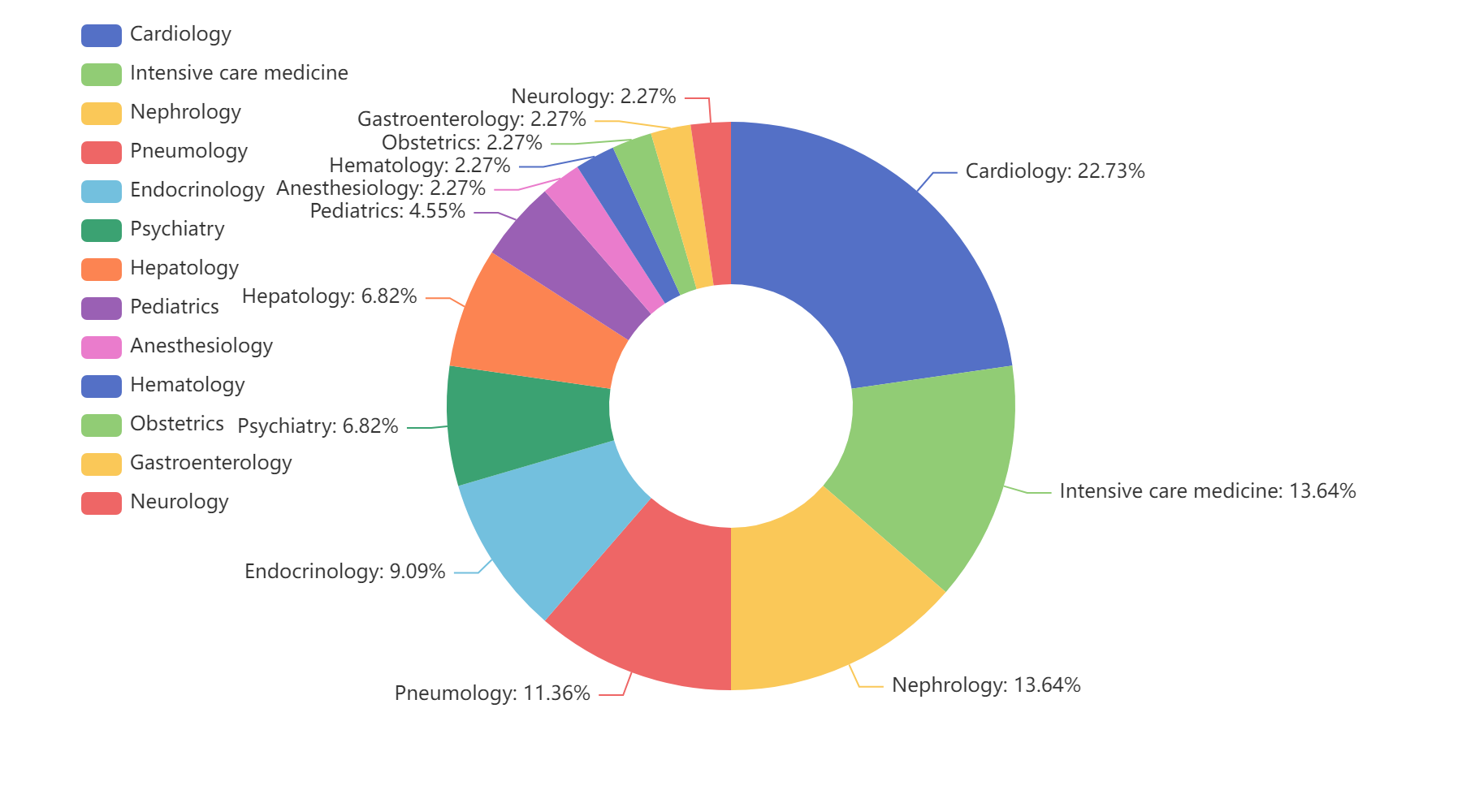}
  \caption{Distribution map of the assessment calculator departments.}
  \label{pie_full}
\end{figure*}

\textbf{Toolkit.} 
Figure \ref{pie_full} presents a detailed distribution of the Medical Assessment Calculators. The calculators we have compiled demonstrate high practicality in clinical practice, especially in 13 critical specialties including Cardiology, Intensive Care Medicine, and Nephrology. These specialties are characterized by a multitude of test items and rely heavily on computational tools to enhance diagnostic precision. This collection of calculator tools is highly reusable and designed to provide strong support and assistance for a broad range of medical assessment activities, thereby optimizing the clinical decision-making process and enhancing the quality of healthcare services. The code for this toolkit is generated by GPT-4 (\citealp{achiam2023gpt}) and has been manually validated.

\textbf{LLM Used.}
All other components of MeNTi uniformly utilize the PULSE model (\citealp{pulse2023}) as their core computational unit. The PULSE model, optimized specifically for the medical field, is an LLM that has demonstrated superior performance and expertise in medical-related tasks. The infrastructure of MeNTi demonstrates a high degree of generalization allowing us to replace the core backbone with other LLMs. In our experiments, we switch to GPT-3.5-turbo and GPT-4o, achieving outstanding performance.

\textbf{Dense Retrieve Used.}
In the meta-tool of the MeNTi system, the Retrieval component employs the Dense Retrieve method for tool selection. Specifically, we use the highly acclaimed M3E retrieval model(\citealp{Moka}). This model is widely recognized for its outstanding performance in semantic representation and similarity calculation.

\textbf{Annotation Details.}
The cases in the annotations all originate from China. Each question of CalcQA required approximately 20 minutes to annotate and review. The cost is \$40 per hour for the physicians.

    

\section{Prompts}\label{apdx: prompt}

\textbf{Toolkit Code Generation Prompt}
\begin{lstlisting}[breaklines=true, breakatwhitespace=true, basicstyle=\small\ttfamily, columns=fullflexible]
You are a code generation model for calculations. You will receive the name, description, and calculation formula of a medical calculator. Your task is to generate a Python function to calculate this calculator based on the provided medical calculator information. 

The requirements are as follows:
1. Your variable names should be as generic and meaningful as possible for easy understanding.
2. You need to provide a docstring that introduces the function's functionality, uses, application scenarios, etc.
3. Your docstring should include detailed descriptions of the function's parameters, their types, and the rules for populating them, to ensure the function can be used correctly.
4. To enhance versatility, if any of the parameters are of string type, you should provide a list of possible types and change the parameter to the index of that list.

The function should be wrapped by
```python
```

Begin!

calculator name: {INSERT_NAME_HERE}
description of calculator: {INSERT_DESCRI_HERE}
formula of calculator: {INSERT_FORMULA_HERE}
\end{lstlisting}

\textbf{Meta-Tool Diagnosis Prompt}
\begin{lstlisting}[breaklines=true, breakatwhitespace=true, basicstyle=\small\ttfamily, columns=fullflexible]
You are a medical diagnostic model. Your task is to analyze the abnormal parts of the provided patient's case and speculate on which bodily functions might be impaired.

The requirements are as follows:
1. Every inference you make must be substantiated by actual evidence from the provided patient's case.
2. You only need to analyze the main, abnormal parts of the provided patient's case.
3. You just need to make a brief analysis.

Begin!
\end{lstlisting}

\textbf{Meta-Tool Classifier Prompt}
\begin{lstlisting}[breaklines=true, breakatwhitespace=true, basicstyle=\small\ttfamily, columns=fullflexible]
You are a toolkit selection model. Below is a toolkit list and their descriptions, and you need to select the appropriate toolkit based on the user query. Your answer should be wrapped by ```json and ```

toolkit list: ["unit", "scale"]
"unit": This is a unit toolkit that contains a variety of medical unit conversion tools. When you need to perform unit conversions, you will need to select this tookit.
"scale": This is a medical calculator toolkit, which is used for assessing and quantifying specific health conditions of individuals in the medical field. When you need to analyze a patient's health condition, or when a user has made a query regarding health status, you will need to select this toolkit.

Requirements:
1. You can only select one toolkit, and it must be from the toolkit list provided.
2. You need to output a JSON file.
3. Your answer should be wrapped by ```json and ```

Please follow this output format:
```json
{
    "chosen_toolkit_name": Str(toolkit you choose)
}
```

Begin!

user query: INSERT_QUERY_HERE
\end{lstlisting}

\textbf{Meta-Tool Rewriter Prompt}
\begin{lstlisting}[breaklines=true, breakatwhitespace=true, basicstyle=\small\ttfamily, columns=fullflexible]
You are a retrieval-augmented model for rewriting queries. You will receive a query from a doctor and a patient's case analysis. Your task is to combine the patient's case analysis to expand and rewrite the doctor's input query, making the doctor's query more aligned with the patient's actual situation.

The requirements are as follows:
1. The generated queries must not alter the doctor's original intent.
2. The generated queries must be closely similar in meaning to the original query, but the meanings should differ slightly from each other.
3. You should extract insights from the patient case analysis that may be related to the doctor's query to generate new queries, in order to facilitate the retrieval of more information. 
4. However, please prioritize the original query; the additional information in each generated query should not be too much to avoid obscuring the content of the original query.
5. You need to generate 3 new queries, neither more nor less.
6. You need to output a JSON file, which is a list where each item is a new query you have generated.
7. You need to answer in English. Your answer should be wrapped by ```json and ```

Please follow this output format:
```json
[
    "the first generated query",
    "the second generated query",
    ...
]
```

Begin!

doctor input search query: INSERT_QUERY_HERE

Patient Case Analysis:

INSERT_CASE_HERE
\end{lstlisting}

\textbf{Meta-Tool Dispatcher Prompt}
\begin{lstlisting}[breaklines=true, breakatwhitespace=true, basicstyle=\small\ttfamily, columns=fullflexible]
You are a dispatching model. Your task is to choose the most suitable tool from the tool list based on User Demand and the Task Scenario, which will then be provided to the user for use.

Tool List: {{INSERT_TOOLLIST_HERE}}
Detailed information of each tool: {{INSERT_TOOLINST_HERE}}

Requirements:
1. You need to conduct a detailed, step-by-step analysis.
2. You must choose a tool from the Tool List.
3. The Final Answer is a JSON file, and the JSON file must be wrapped by ```json and ```
4. The tool you choose in the JSON file must be one of the items in the Tool List.

Here is a example of the Final Answer:
```json
{
    "chosen_tool_name": Str(the tool you choose)
}
```

Begin!

User Demand: {{INSERT_DEMAND_HERE}}
Task Scenario: {{INSERT_SCE_HERE}}
Step By Step Analysis:
\end{lstlisting}

\textbf{Slot Filling Prompt}
\begin{lstlisting}[breaklines=true, breakatwhitespace=true, basicstyle=\small\ttfamily, columns=fullflexible]
You are a parameter extraction model. You will receive a Reference Text and a Function Docstring. Your task is to determine the parameters from the Reference Text based on the parameter filling rules described in the Function Docstring, including the values and units of the parameters.

The requirements are as follows:
1. The Value and Unit of parameters you output need to be strictly in accordance with the Reference Text. You are prohibited from performing unit conversions.
2. If there is a discrepancy in the unit of the parameter between the Reference Text and the Function Docstring, please use the unit from the Reference Text as the standard. Do not convert the units on your own.
3. If the parameter does not have a unit, output 'null' in the Unit.
4. All parameters in the Function Docstring must be included in the parameter list. If the parameter values are missing, fill them randomly. The Value must not be 'null'.
5. For parameters that do not have a clear rating in the Reference Text, please infer and fill them out based on the actual circumstances described in the reference text and the scoring standards provided in the Function Docstring.
6. You need to first produce a step-by-step analysis, considering each parameter individually.
7. The Parameters List you output is a JSON file, and this JSON file should be wrapped by ```json and ```

Please follow this output format:
```json
{The parameters list here.}
```

Here are some examples:
Function Docstring: 
{{"Calculate the Body Mass Index (BMI) for an individual.\n\nArgs:\nweight (float): The weight of the individual in kilograms.\nheight (float): The height of the individual in centimeters.\n\nReturns:\nfloat: the BMI (kg/m^2).\n\nDescription:\nThe Body Mass Index (BMI) is a simple index of weight-for-height commonly used to classify\nunderweight, overweight, and obesity in adults. It is calculated by dividing the weight in\nkilograms by the square of the height in meters. Although widely used, BMI has limitations,\nparticularly for very muscular individuals and in different ethnic groups with varying body\nstatures, where it may not accurately reflect body fat percentages."}}
Reference Text:
{{The patient is a 16-year-old male, 175cm in height and 65kg in weight}}
Step By Step Analysis:
{{Here is your step-by-step analysis.}}
Parameters List:
```json
{
    "weight": {"Value": 65, "Unit": "kg"},
    "height": {"Value": 175, "Unit": "cm"}
}
```

Begin!

Function Docstring: 
{{INSERT_DOCSTRING_HERE}}
Reference Text:
{{INSERT_TEXT_HERE}}
Step By Step Analysis:
\end{lstlisting}

\textbf{Nested Calling Prompt}
\begin{lstlisting}[breaklines=true, breakatwhitespace=true, basicstyle=\small\ttfamily, columns=fullflexible]
You are a Parameter List checking model.

You will receive a Function Docstring, and Parameter List. You need to verify that the entries in the Parameter List comply with the requirements described in the Function Docstring, including the Value and Unit.
If all units are consistent, choose "calculate". If there are any discrepancies in the units, choose "toolcall".
You should not perform unit conversions directly. When converting units is needed, you must choose "toolcall" and elaborate on this unit conversion task in the "supplementary_information", including the parameter value, the current unit of the parameter, and the target unit of the parameter.

Requirements:
1. You need to conduct a detailed, step-by-step analysis of each parameter in the parameter list. You need to output each parameter's Function Docstring individually, then analyze and compare them.
2. You especially need to compare the Unit of the Parameter List with the units required in the Function Docstring. 
3. If the units are inconsistent, please select "toolcall" and specify the numerical value of the parameter required for unit conversion, as well as the units before and after the conversion in the "supplementary_information".
4. The unit conversion task may require converting units of different parameters, and you need to break down the task into individual unit conversion tasks for each parameter. Therefore, "supplementary_information" is a list of strings, each of which represents a standalone, minimalized unit conversion task.
5. The Final Answer is a JSON file, and the JSON file must be wrapped by ```json and ```

Here are some examples:
Function Docstring:
{{"Calculate the Body Mass Index (BMI) for an individual.\n\nArgs:\nweight (float): The weight of the individual in kilograms.\nheight (float): The height of the individual in centimeters.\n\nReturns:\nfloat: the BMI (kg/m^2).\n\nDescription:\nThe Body Mass Index (BMI) is a simple index of weight-for-height commonly used to classify\nunderweight, overweight, and obesity in adults. It is calculated by dividing the weight in\nkilograms by the square of the height in meters. Although widely used, BMI has limitations,\nparticularly for very muscular individuals and in different ethnic groups with varying body\nstatures, where it may not accurately reflect body fat percentages."}}
Parameter List:
{{{
    "weight": {"Value": 65, "Unit": "kg"},
    "height": {"Value": 1.75, "Unit": "m"}
}}}
Step By Step Analysis:
{{(Here is your step-by-step analysis. You need to ouptput the corresponding Function Docstring first for each parameter, and systematically compare each parameter with the corresponding information in the Parameter List and Function Docstring.)}}
Final Answer:
```json
{
    "chosen_decision_name": "toolcall",
    "supplementary_information": ["The height is 1.75m. The height needs to be converted from meters to centimeters."]
}
```

Function Docstring:
{{Calculate the Corrected Sodium for Hyperglycemia using Hillier's method from 1999.\n\nParameters:\n measured_sodium (float): The measured sodium level in mEq/L.\n serum_glucose (float): The serum glucose level in mg/dL.\n\nReturns:\n float: The corrected sodium level in mEq/L.\n\n}}
Parameter List:
{{{
    "measured_sodium": {"Value": 140, "Unit": "mmol/L"},
    "serum_glucose": {"Value": 80, "Unit": "mmol/L"}
}}}
Step By Step Analysis:
{{(Here is your step-by-step analysis. You need to ouptput the corresponding Function Docstring first for each parameter, and systematically compare each parameter with the corresponding information in the Parameter List and Function Docstring.)}}
Final Answer:
```json
{
    "chosen_decision_name": "toolcall",
    "supplementary_information": ["The measured_sodium is 140 mmol/L. It needs to be converted from mmol/L to mEq/L.", "The serum_glucose is 80 mmol/L. It needs to be converted from mmol/L to mg/dL."]
}
```

Function Docstring:
{{"Calculate the Body Mass Index (BMI) for an individual.\n\nArgs:\nweight (float): The weight of the individual in kilograms.\nheight (float): The height of the individual in centimeters.\n\nReturns:\nfloat: the BMI (kg/m^2).\n\nDescription:\nThe Body Mass Index (BMI) is a simple index of weight-for-height commonly used to classify\nunderweight, overweight, and obesity in adults. It is calculated by dividing the weight in\nkilograms by the square of the height in meters. Although widely used, BMI has limitations,\nparticularly for very muscular individuals and in different ethnic groups with varying body\nstatures, where it may not accurately reflect body fat percentages."}}
Parameter List:
{{{
    "weight": {"Value": 65, "Unit": "kg"},
    "height": {"Value": 175, "Unit": "cm"}
}}}
Step By Step Analysis:
{{(Here is your step-by-step analysis. You need to ouptput the corresponding Function Docstring first for each parameter, and systematically compare each parameter with the corresponding information in the Parameter List and Function Docstring.)}}
Final Answer:
```json
{
    "chosen_decision_name": "calculate",
    "supplementary_information": "All parameters comply with the Function Docstring requirements. No unit conversion is needed as the parameters use indices to specify units."
}
```

Attention: The Final Answer must be wrapped by ```json and ```. In the 'supplementary_information', the Value of the parameter is required!!!

Begin!

Function Docstring:
{{INSERT_DOC_HERE}}
Parameter List: {{INSERT_LIST_HERE}}
Step By Step Analysis:
\end{lstlisting}

\section{Toolkit Example}
In our toolkit, each tool is meticulously integrated with its core information, encompassing the tool name, function name, description, code, and accompanying code docstring. Depending on the tool's classification and the diversity of the sources of information, we have further supplemented specific information. For instance, for tools of scale assessments, given the complexity and significance of their computational methods, a dedicated column for calculation formulas has been added to ensure users can accurately understand the algorithmic logic. To use these tools, simply follow the clear instructions provided in the function docstrings to configure the necessary parameters, and then directly invoke the provided code to execute the functions. Finally get the value of calculation results. This process is both intuitive and efficient. Below is an example of a typical tool for reference:

\begin{lstlisting}[breaklines=true, breakatwhitespace=true, basicstyle=\small\ttfamily, columns=fullflexible]
{
    "tool_name": "Framingham Risk Score for Hard Coronary Heart Disease",
    "function_name": "calculate_framingham_risk_score",
    "description": "The Framingham Risk Score for Hard Coronary Heart Disease is a tool used by healthcare professionals to estimate a patient's 10-year risk of developing severe heart disease. It utilizes factors such as age, gender, cholesterol levels, blood pressure, and smoking status to calculate risk. This score helps in identifying individuals at high risk, guiding decisions on preventive measures and interventions.",
    "formula": "Here are the parameter list: \n\nAge: years\nSex: Female/Male\n Smoker: No/Yes\nTotal cholesterol: mg/dL\nHDL cholesterol: mg/dL\nSystolic BP: mm Hg\nBlood pressure being treated with medicines: No/Yes\n\nEquations as follows:\n\nMen:\nLMen = 52.00961 x ln(Age) + 20.014077 x ln(Total cholesterol) - 0.905964 x ln(HDL cholesterol) + 1.305784 x ln(Systolic BP) + 0.241549 x Treated for blood pressure + 12.096316 x Smoker - 4.605038 x ln(Age) x ln(Total cholesterol) - 2.84367 x ln(Age) x Smoker - 2.93323 x ln(Age) x ln(Age) - 172.300168\nPMen = 1 - 0.9402^exp(LMen)\n\nWomen:\nLWomen = 31.764001 x ln(Age) + 22.465206 x ln(Total cholesterol) - 1.187731 x ln(HDL cholesterol) + 2.552905 x ln(Systolic BP) + 0.420251 x Treated for blood pressure + 13.07543 x Smoker - 5.060998 x ln(Age) x ln(Total cholesterol) - 2.996945 x ln(Age) x Smoker - 146.5933061\nPWomen = 1 - 0.98767^exp(LWomen)\n\nSupplementary item:\n1. Smoker: Yes = 1, No = 0.\n2. Men: if age >70, use ln(70) x Smoker. Women: if age >78, use ln(78) x Smoker.\n",
    "code": "import math\n\n\ndef calculate_framingham_risk_score(age, sex, smoker_status, total_cholesterol, hdl_cholesterol, systolic_bp,\n                                    bp_medication):\n    \"\"\"\n    Calculate the Framingham Risk Score for Hard Coronary Heart Disease in patients aged 30-79 years without prior coronary heart disease history, diabetes, or intermittent claudication.\n\n    Parameters:\n    - age (int): The age of the patient in years. Valid range: 30-79.\n    - sex (int): The sex of the patient. 0 for female, 1 for male.\n    - smoker_status (int): Smoking status of the patient. 0 for non-smoker, 1 for smoker.\n    - total_cholesterol (float): Total cholesterol level in mg/dL.\n    - hdl_cholesterol (float): HDL cholesterol level in mg/dL.\n    - systolic_bp (float): Systolic blood pressure in mm Hg.\n    - bp_medication (int): Indicates if the blood pressure is being treated with medications. 0 for no, 1 for yes.\n\n    Returns:\n    float: The risk percentage of developing hard coronary heart disease.\n\n    Calculation is based on the logarithmic transformations of risk factors and their interactions,\n    separately for males and females. Note that older population data was used to develop this scale,\n    which may not fully align with current population risks.\n    \"\"\"\n    # Converting inputs for equation\n    ln_age = math.log(age)\n    ln_total_cholesterol = math.log(total_cholesterol)\n    ln_hdl_cholesterol = math.log(hdl_cholesterol)\n    ln_systolic_bp = math.log(systolic_bp)\n\n    if sex == 0:  # Female\n        if age > 78:\n            age_smoker_interaction = math.log(78) * smoker_status\n        else:\n            age_smoker_interaction = ln_age * smoker_status\n\n        L = (31.764001 * ln_age +\n             22.465206 * ln_total_cholesterol -\n             1.187731 * ln_hdl_cholesterol +\n             2.552905 * ln_systolic_bp +\n             0.420251 * bp_medication +\n             13.07543 * smoker_status -\n             5.060998 * ln_age * ln_total_cholesterol -\n             2.996945 * age_smoker_interaction -\n             146.5933061)\n\n        P = 1 - 0.98767 ** math.exp(L)\n\n    else:  # Male\n        if age > 70:\n            age_smoker_interaction = math.log(70) * smoker_status\n        else:\n            age_smoker_interaction = ln_age * smoker_status\n\n        L = (52.00961 * ln_age +\n             20.014077 * ln_total_cholesterol -\n             0.905964 * ln_hdl_cholesterol +\n             1.305784 * ln_systolic_bp +\n             0.241549 * bp_medication +\n             12.096316 * smoker_status -\n             4.605038 * ln_age * ln_total_cholesterol -\n             2.84367 * age_smoker_interaction -\n             2.93323 * ln_age ** 2 -\n             172.300168)\n\n        P = 1 - 0.9402 ** math.exp(L)\n\n    return P * 100  # Convert to percentage\n",
    "docstring": "Calculate the Framingham Risk Score for Hard Coronary Heart Disease in patients aged 30-79 years without prior coronary heart disease history, diabetes, or intermittent claudication.\n\n    Parameters:\n    - age (int): The age of the patient in years. Valid range: 30-79.\n    - sex (int): The sex of the patient. 0 for female, 1 for male.\n    - smoker_status (int): Smoking status of the patient. 0 for non-smoker, 1 for smoker.\n    - total_cholesterol (float): Total cholesterol level in mg/dL.\n    - hdl_cholesterol (float): HDL cholesterol level in mg/dL.\n    - systolic_bp (float): Systolic blood pressure in mm Hg.\n    - bp_medication (int): Indicates if the blood pressure is being treated with medications. 0 for no, 1 for yes.\n\n    Returns:\n    float: The risk percentage of developing hard coronary heart disease.\n\n    Calculation is based on the logarithmic transformations of risk factors and their interactions,\n    separately for males and females. Note that older population data was used to develop this scale,\n    which may not fully align with current population risks.\n"
}
\end{lstlisting}

\section{Inference Example}\label{apdx: example}

\begin{tcolorbox}[title={User Query}, colframe=color_consultation, breakable]
What scale should be used to assess a patient's risk of Coronary heart attack?
\tcblower
Basic information: male, 49, civil servants

Chief Complaints: Chest tightness and shortness of breath January

History of present disease:
1 month ago, there was no incentive for chest tightness and asthma, mostly at night, each lasting about 1 hour, can be alleviated by itself, no dizziness, headache, syncope, dark day, nausea, vomiting, cough, phlegm, palpitations, abdominal pain, diarrhea, edema of both lower limbs and other discomfort. Chest CT showed: a high high-density shadow of two upper lung apexes and pleural effusion on both sides. B-ultrasonography showed bilateral pleural effusion. The ECG showed sinus rhythm, left ventricular hypertrophy, left atrial load increase, and some lead T-wave changes. Color Doppler echocardiography indicated that the left heart was enlarged and the ejection fraction of the left heart was decreased. The symptoms were not alleviated significantly after drug treatment (specific details are unknown). Coronary angiography was recommended, and the patient was hospitalized in our hospital. During this period, the patient's mental appetite and sleep are OK, and urine and bowel have no obvious abnormalities.

Previous history:
The patient was found to have elevated blood pressure for 5 years, with a maximum blood pressure of 180/100 MMHG, taking oral antihypertensive drugs and monitoring blood pressure. History of diabetes 3~4 years, oral metformin tablets 0.5g, blood sugar control is good. A history of smoking. The patient's mother had a history of diabetes, and his father had a history of hypertension and coronary heart disease.

Physical Examination:
T: 36.5℃, P: 107 times/min, R: 18 times/min, BP: 160/110mmHg
God clear, eyelid no edema, sclera no yellow staining, soft neck, jugular vein no angry expansion, liver jugular reflux sign negative, thyroid gland no swelling. The trachea was centered, the chest was not malformed, the respiratory sounds of the two lower lungs were slightly lower, and the dry and wet rales were not heard, and there was no pleural friction sound. There was no abnormal eminence in the precardiac area and no uplifting beat. The apex beat was in the fifth intercostal space above the left midclavicular line, and the cardiac boundary expanded to the left lower. The rhythm was 107 beats/min, and the rhythm was uniform. The whole abdomen was soft, without tenderness and rebound pain, the liver, spleen and ribs were not touched, both kidneys were not touched, the mobile dullness was negative, and the intestinal ringing was 4-5 times/min. There was no edema in both lower limbs. Physiological reflex was present, but pathological reflex was not induced.

Auxiliary Examination:
Blood routine, liver and kidney function, electrolyte, thyroid function, troponin, creatine kinase isoenzyme, and tumor markers were not abnormal. Blood biochemical test: total cholesterol: 8.3mmol/L, high-density lipoprotein cholesterol: 0.2mmol/L, low-density lipoprotein cholesterol (LDL-C) 4.1mmol/L brain natriuretic peptide (NT-proBNP) 1013 ng/L. The results of B-ultrasound showed that fatty liver, biliary pancreas, and spleen were not abnormal. Cardiac color ultrasonography showed left atrial and left ventricular enlargement [left atrial diameter (LAD) 50 mm; left ventricular systolic diameter (LVD) 56mm; left ventricular diastolic diameter (LVDd) 66 mm], cardiac insufficiency (LVEF 44\%), mild mitral insufficiency, and mild aortic insufficiency. Holter electrocardiogram showed: sinus rhythm, frequent ventricular premature, short ventricular tachycardia, occasionally unsustained atrial tachycardia, intermittent T wave low level. Ambulatory blood pressure: mean blood pressure 150/92 MMHG, maximum blood pressure 185/105 MMHG. Chest CT showed left ventricular enlargement.
\end{tcolorbox}

\begin{tcolorbox}[title={Meta-Tool Classifier}, colframe=color_curation, breakable]
    Use the calculator toolkit.
\end{tcolorbox}

\begin{tcolorbox}[title={Meta-Tool Rewriter}, colframe=color_curation, breakable]
\begin{itemize}
    \item What is the best assessment scale for cardiovascular dysfunction, considering the patient's symptoms of chest tightness, shortness of breath, ECG abnormalities, previous hypertension, and reduced ejection fraction?
    \item Which scale should be used to evaluate the risk of a heart attack in a patient with a history of smoking, family history of diabetes and hypertension, and current cardiovascular, respiratory, and metabolic impairments?
    \item What risk assessment method is suitable for a coronary heart attack in a patient with histories of hypertension and diabetes, elevated cholesterol levels, decrease in HDL, and impaired liver function indicated by fatty liver?
\end{itemize}
\end{tcolorbox}

\begin{tcolorbox}[title={Meta-Tool Retrieval}, colframe=color_curation, breakable]
\begin{enumerate}
    \item HEART Score for Major Cardiac Events
    \item Framingham Risk Score for Hard Coronary Heart Disease
    \item Revised Cardiac Risk Index for Pre-Operative Risk
    \item CHA2DS2-VASc Score for Atrial Fibrillation Stroke Risk
    \item Mean Arterial Pressure (MAP)
\end{enumerate}
\end{tcolorbox}

\begin{tcolorbox}[title={Meta-Tool Dispatcher}, colframe=color_curation, breakable]
Step 1: Understanding User Demand
The user demands a tool to assess a patient's risk of a coronary heart attack. Having a high risk of a heart attack could help in early diagnosis and preventive measures. 

Step 2: Analyzing the Task Scenario
The task scenario is a description of a patient suffering from several health issues including hypertension, potential cardiovascular disease, potential respiratory issues, metabolic dysfunction, and potential liver impairment. 

Step 3: Matching User Demand and Task Scenario to a Tool
Comparing the user's requirement and the case, the tool needed is one that can assess the risk of coronary heart disease given the patient's condition, including multiple cardiovascular risk factors, such as diabetes, hypertension, elevated cholesterol levels, and smoking history. 

Step 4: Choosing the Most Suitable Tool
Based on the user's requirement and the task scenario, the Framingham Risk Score for Hard Coronary Heart Disease would be the most suitable tool. This tool helps to evaluate the risk of coronary heart disease in patients without a prior history of the disease. It considers variables such as age, sex, smoking status, total cholesterol, HDL cholesterol, systolic blood pressure, and blood pressure treatment, which would accurately reflect the patient's medical history and current condition. 

Framingham Risk Score for Hard Coronary Heart Disease
\end{tcolorbox}

\begin{tcolorbox}[title={Slot Filling}, colframe=color_configuration, breakable]
\begin{lstlisting}[breaklines=true, breakatwhitespace=true, basicstyle=\small\ttfamily, columns=fullflexible]
{
    "age": {"Value": 49, "Unit": "years"},
    "sex": {"Value": 1, "Unit": "null"},
    "smoker_status": {"Value": 1, "Unit": "null"},
    "total_cholesterol": {"Value": 8.3, "Unit": "mmol/L"},
    "hdl_cholesterol": {"Value": 0.2, "Unit": "mmol/L"},
    "systolic_bp": {"Value": 160, "Unit": "mmHg"},
    "bp_medication": {"Value": 1, "Unit": "null"}
}
\end{lstlisting}
\end{tcolorbox}

\begin{tcolorbox}[title={Nested Calling}, colframe=color_configuration, breakable]
\begin{lstlisting}[breaklines=true, breakatwhitespace=true, basicstyle=\small\ttfamily, columns=fullflexible]
{
    "chosen_decision_name": "toolcall",
    "supplementary_information": [
        "The total_cholesterol is 8.3 mmol/L. It needs to be converted from mmol/L to mg/dL.",
        "The hdl_cholesterol is 0.2 mmol/L. It needs to be converted from mmol/L to mg/dL."
    ]
}
\end{lstlisting}
\end{tcolorbox}

\begin{tcolorbox}[title={MeNTi}, colframe=color_consultation, breakable]
The total\_cholesterol is 8.3 mmol/L. It needs to be converted from mmol/L to mg/dL.
\end{tcolorbox}

\begin{tcolorbox}[title={Meta-Tool Rewriter}, colframe=color_curation, breakable]
\begin{itemize}
    \item "How to convert 8.3 mmol/L total cholesterol to mg/dL?",
    \item "Guidelines for conversion of total cholesterol from mmol/L to mg/dL",
    \item "Can I convert 8.3 mmol/L total cholesterol level to mg/dL?",
\end{itemize}
\end{tcolorbox}

\begin{tcolorbox}[title={Meta-Tool Retrieval}, colframe=color_curation, breakable]
\begin{enumerate}
    \item Total Cholesterol
    \item Low-density lipoprotein cholesterol
    \item High-density lipoprotein cholesterol
    \item Ethylene glycol
    \item Methanol
\end{enumerate}
\end{tcolorbox}

\begin{tcolorbox}[title={Meta-Tool Dispatcher}, colframe=color_curation, breakable]
Total Cholesterol.
\end{tcolorbox}

\begin{tcolorbox}[title={Slot Filling}, colframe=color_configuration, breakable]
\begin{lstlisting}[breaklines=true, breakatwhitespace=true, basicstyle=\small\ttfamily, columns=fullflexible]
{
    "input_value": {"Value": 8.3, "Unit": "null"},
    "input_unit": {"Value": 0, "Unit": "null"},
    "target_unit": {"Value": 2, "Unit": "null"}
}
\end{lstlisting}
\end{tcolorbox}

\begin{tcolorbox}[title={Nested Calling}, colframe=color_configuration, breakable]
\begin{lstlisting}[breaklines=true, breakatwhitespace=true, basicstyle=\small\ttfamily, columns=fullflexible]
{
  "chosen_decision_name": "calculate",
  "supplementary_information": null
}

For the Total Cholesterol, 8.3 mmol/L is equal to 320.9195 mg/dL
\end{lstlisting}
\end{tcolorbox}

\begin{tcolorbox}[title={MeNTi}, colframe=color_consultation, breakable]
The hdl\_cholesterol is 0.2 mmol/L. It needs to be converted from mmol/L to mg/dL.
\end{tcolorbox}

\begin{tcolorbox}[title={Meta-Tool Rewriter}, colframe=color_curation, breakable]
\begin{itemize}
    \item "How to convert the HDL cholesterol level from mmol/L to mg/dL when the value is 0.2",
    \item "Conversion of 0.2 mmol/L HDL cholesterol to mg/dL",
    \item "What is 0.2 mmol/L of HDL cholesterol in mg/dL?",
\end{itemize}
\end{tcolorbox}

\begin{tcolorbox}[title={Meta-Tool Retrieval}, colframe=color_curation, breakable]
\begin{enumerate}
    \item Total Cholesterol
    \item High-density lipoprotein cholesterol
    \item Low-density lipoprotein cholesterol
\end{enumerate}
\end{tcolorbox}

\begin{tcolorbox}[title={Meta-Tool Dispatcher}, colframe=color_curation, breakable]
High-density lipoprotein cholesterol
\end{tcolorbox}

\begin{tcolorbox}[title={Slot Filling}, colframe=color_configuration, breakable]
\begin{lstlisting}[breaklines=true, breakatwhitespace=true, basicstyle=\small\ttfamily, columns=fullflexible]
{
    "input_value": {"Value": 0.2, "Unit": "mmol/L"},
    "input_unit": {"Value": 0, "Unit": null},
    "target_unit": {"Value": 2, "Unit": null}
}
\end{lstlisting}
\end{tcolorbox}

\begin{tcolorbox}[title={Nested Calling}, colframe=color_configuration, breakable]
\begin{lstlisting}[breaklines=true, breakatwhitespace=true, basicstyle=\small\ttfamily, columns=fullflexible]
{
  "chosen_decision_name": "calculate",
  "supplementary_information": null
}

For the High-density lipoprotein cholesterol, 0.2 mmol/L is equal to 7.7330000000000005 mg/dL
\end{lstlisting}
\end{tcolorbox}

\begin{tcolorbox}[title={Slot Filling}, colframe=color_configuration, breakable]
\begin{lstlisting}[breaklines=true, breakatwhitespace=true, basicstyle=\small\ttfamily, columns=fullflexible]
{
    "age": {"Value": 49, "Unit": "years"},
    "sex": {"Value": 1, "Unit": "null"},
    "smoker_status": {"Value": 1, "Unit": "null"},
    "total_cholesterol": {"Value": 320.9195, "Unit": "mg/dL"},
    "hdl_cholesterol": {"Value": 7.733, "Unit": "mg/dL"},
    "systolic_bp": {"Value": 160, "Unit": "mmHg"},
    "bp_medication": {"Value": 1, "Unit": "null"}
}
\end{lstlisting}
\end{tcolorbox}

\begin{tcolorbox}[title={Nested Calling}, colframe=color_configuration, breakable]
\begin{lstlisting}[breaklines=true, breakatwhitespace=true, basicstyle=\small\ttfamily, columns=fullflexible]
{
    "chosen_decision_name": "calculate",
    "supplementary_information": "All parameters comply with the Function Docstring requirements. No unit conversion is needed as the parameters use correct units or indices."
}
\end{lstlisting}
\end{tcolorbox}

\begin{tcolorbox}[title={MeNTi}, colframe=color_consultation, breakable]
The patient's Framingham Risk Score for Hard Coronary Heart Disease is 93.70109147053569
\end{tcolorbox}


\end{document}